\pdfoutput=1
\documentclass[11pt]{article}

\usepackage{acl}

\usepackage{times}
\usepackage{latexsym}

\usepackage[T1]{fontenc}
\usepackage[utf8]{inputenc}

\usepackage{microtype}

\usepackage{graphicx}
\usepackage{algorithm, amsmath, amsfonts, amsthm, mleftright, multirow}
\usepackage{siunitx}
\hypersetup{breaklinks=true}

\usepackage{caption}

\theoremstyle{definition}

\usepackage{algpseudocode}
\usepackage{fvextra}

\usepackage{microtype}

\title{Improving Few-Shot Image Classification Using Machine- and User-Generated Natural Language Descriptions}
\author{Kosuke Nishida,
		Kyosuke Nishida,
		Shuichi Nishioka\\
		\rm NTT Human Informatics Laboratories, NTT Corporation\\
		\tt kosuke.nishida.ap@hco.ntt.co.jp}

\date{}

\begin{document}
\maketitle
\begin{abstract}
Humans can obtain the knowledge of novel visual concepts from language descriptions, and we thus use the few-shot image classification task to investigate whether a machine learning model can have this capability. Our proposed model, LIDE (Learning from Image and DEscription), has a text decoder to generate the descriptions and a text encoder to obtain the text representations of machine- or user-generated descriptions. We confirmed that LIDE with machine-generated descriptions outperformed baseline models. Moreover, the performance was improved further with high-quality user-generated descriptions. The generated descriptions can be viewed as the explanations of the model's predictions, and we observed that such explanations were consistent with prediction results. We also investigated why the language description improved the few-shot image classification performance by comparing the image representations and the text representations in the feature spaces.
	%We also analyzed the manifolds of the image features and the multi-modal features to clarify why LIDE improved the image classification performance.
\end{abstract}

\section{Introduction}
Humans can efficiently learn about new concepts from language \cite{concept}. 
%By virtue of large-scale pre-training, machine learning models are also acquiring the capability to learn from natural language descriptions \cite{gpt-2, gpt-3, flan}.
Hence, in this paper, we focus on the few-shot image classification problem to verify machine learning models' capability to understand new concepts from language. This problem is a kind of meta-learning problem in which a model first learns from the concepts of classes 
by training on a few instances and then learns unseen classes in the same way. 
%Each learning phase has support instances for learning and query instances for evaluation. %consists of training and test.

In our problem setting, the model can use language descriptions of images as additional information.
%both during training and inference.
%even if there is no ground-truth description in the test time.
%As a new problem setting, we provide the language description of the image as the additional information.
%The model can learn new concepts from both support images and texts, and then it classifies an image based solely on its content or on both the image and text.
%to the unseen classes.
This setting is similar to teaching a new concept to others by explaining it deductively from small amounts of data, unlike most machine learning models that learn inductively from large amounts of data.
%\textcolor{blue}{From another viewpoint of vision-and-language, this setting contributes to the evaluation of human-like fast learning of unseen image classes by using the capability of visually grounded language understanding.}

%When handling with unseen classes during inference, this setting allows the model to use the text describing the class or instance as support.
%言語から新しいコンセプトが学習できる->ので，推論時に言語情報からコンセプトを学習できる->言語情報がサポートとして使える　というロジックです
%Therefore, we can  through few person teaching a new concept to oth-shot image classification.
% In this paper, we guide visual representation learning with language, studying the setting where no language is available at test time, since rich linguistic supervision is often unavailable for new concepts encountered in the wild. 

\begin{figure}
		\includegraphics[width =75mm]{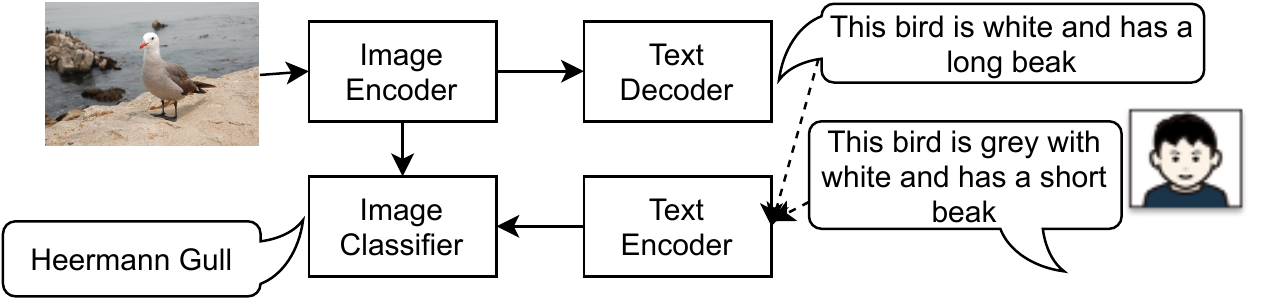}
		\caption{Concept of the LIDE model.} %The system in \citet{lsl} does not have the text encoder and use the text decoder as the regularizer in the training time.}
		\label{fig:concept}
\end{figure}

For this setting, we propose a new model, called \textit{LIDE (Learning from Image and DEscription)}.
%}, that learns from images and natural language descriptions.
As shown in Figure~\ref{fig:concept}, LIDE consists of an image encoder, an image classifier, a text decoder, and a text encoder.

LIDE has the advantage of providing explainablity. It passes an image representation encoded by the image encoder to the text decoder, which then generates a language description of the image as an explanation of the model's prediction. The image classifier then classifies the image in accordance with the text representation, which is encoded by the text encoder, in addition to the image representation.

LIDE also provides high accuracy due to its explainability. It has been difficult to use machine-generated descriptions to improve image classification performance because of their low quality~\cite{lsl}. Therefore, we design a training algorithm and a text encoding method to obtain robust text representations. In addition, LIDE includes a feature fusion module added to the image classifier to combine the information from both the image representation and the text representation.
%However, \citet{lsl} reported that is is difficult to use machine-generated descriptions to improve the image classification performance because of their low quality. %~\cite{lsl}.

%To encode such descriptions robustly, we introduce new techniques to enrich the encoded text representations: The contrastive learning between the gold captions and generated captions to make them close, and the weighted average pooling of the encoded token representation to ignore low-confidence tokens.
%but LIDE can reduce the influence of the low-quality outputs of the text decoder by introducing a weighted average pooling based on the confidence of each token in the generated description. 
%As a result, LIDE improves the image classification accuracy by combining the information from both the image representation and the text representation.

Moreover, LIDE can take user-generated descriptions as input instead of machine-generated descriptions. 
%\textcolor{blue}{Such user-generated descriptions enable the use of features captured from an input image by human perception or from a post-edited text in the text decoder output}. 
We can use a description that contains textual features captured from an input image by human perception or a post-edited text in the text decoder output. % as the user-generated descriptions.
The resulting high-quality descriptions provided by users can improve the image classification accuracy.

Our contributions are summarized as follows:

\begin{itemize}
    \item We confirmed that LIDE with machine-generated descriptions outperformed previous models,
    %using only the image representations.
    %We also found that 
    and thus the explanations of the model's predictions were helpful to improve the classification accuracy.
    
    \item We observed that the performance improved further when 
    gold captions were fed to LIDE as users' high-quality descriptions. 
    %were treated as gold captions.}
    % users' high-quality descriptions were available by viewing the gold capions as them.
    
    \item We investigated whether the generated explanations were consistent with the image classification predictions,
    and we found a positive correlation between the quality of the generated captions and the classification accuracy.
    
    \item We thoroughly investigated why the text representations explaining the input image contributed to the image classification task, specifically in terms of the distributions of the representations in the feature space, the robustness of the representations for noisy images, and the knowledge of concepts that can be extracted from the representations.
    %\textcolor{blue}{some of birds' attributions recovered easily from text representations.} %each representation modality.
\end{itemize}

%In fact, verbal and written language form the basis for much of human learning and pedagogy, as reflected in text-books, lectures and student-teacher dialogues. Natural language explanations can be a potent mode of supervision, and can alleviate issues of data sparsity by directly encoding relevant knowledge about concepts. F

% Humans are powerful and efficient learners partially due to the ability to learn from language (Chopra et al., 2019; Tomasello, 1999). 
%For instance, we can learn about robins not by seeing thousands of examples, but by being told that a robin is a bird with a red belly and brown feathers. 
% This language further shapes the way we view the world, constraining our hypotheses for new concepts: given a new bird (e.g. seagulls), even without language we know that features like belly and feather color are relevant (Goodman, 1955). 

%In this paper, we guide visual representation learning with language, studying the setting where no language is available at test time, since rich linguistic supervision is often unavailable for new concepts encountered in the wild. 

%How can one best use language in this setting?  One option is to just regularize, training representations to predict language descriptions. Another is to exploit the compositional nature of language directly by using it as a bottleneck in a discrete latent variable model.

\section{Background}
\subsection{Few-Shot Image Classification}
$N$-way $K$-shot classification involves three data splits, $\mathcal{T}_{train}, \mathcal{T}_{dev}$, and $\mathcal{T}_{test}$, and each split consists of many classes and instances. The classes in the splits are disjoint. 
%Therefore, different to the standard image classification setting, \textcolor{blue}{a classification model has to recognize the novel concepts in the instances and predict the classification of the instances into unseen classes in the development and test splits.}
%%%
Each $N$-way $K$-shot classification task is a classification problem with $N$ classes. Each task provides $K$ training instances for each class, %which are 
called \textit{support} instances. A task entails the evaluation of $M$ instances for each class, %which are 
called \textit{query} instances. 
%The classes and instances are randomly sampled from \textcolor{blue}{the data splits}.
The tasks, which consist of the classes and instances, are randomly sampled from the data splits.

This problem is a meta-learning problem.
In the training phase, we use the episodic training \cite{episode}, %with 
%the support and query examples of $\mathcal{T}_{train}$ 
%\textcolor{magenta}{multiple tasks sampled from $\mathcal{T}_{train}$  for each mini-batch}
%\textcolor{magenta}{$\mathcal{T}_{train}$
%to train the model,}
%That is, we sampled the batch size number of tasks consisting of $N$ classes and $N(K+M)$ instances from $\mathcal{T}_{train}$ independently for the mini-batch training. 
%\textcolor{red}{That is, tasks consisting of $N$ classes and $N(K+M)$ instances from $\mathcal{T}_{train}$ are sampled independently as a mini-batch. }
where 
%the training dataset are split into small mini-batches, and each mini-batch 
many mini-batches
of size $B$ consisting of $B$ tasks, each of which consists of $N$ classes and $N(K+M)$ instances, are independently sampled from $\mathcal{T}_{train}$. 
%In the test phase, the model learns new concepts from the supports sampled from $\mathcal{T}_{dev}$ $\mathcal{T}_{test}$, and we evaluate the classification performance of the queries.
In the test phase, the model learns new $N$-way $K$-shot classification
tasks with unseen classes and instances sampled from $\mathcal{T}_{dev}$ or $\mathcal{T}_{test}$. For each sampled task, a model learns from the supports, and we evaluate the classification performance of the queries.

A major approach for $N$-way $K$-shot classification is the prototypical network (ProtoNet) \cite{proto}.
In both the training and test phase, instead of updating the model parameters for each sampled task with few support instances, the prototypical network computes the class prototypes. Let $h^c_k$ be the $k$-th support feature of class $c$. Here, the class prototype $z^c$ is
\[
z^c = \frac{1}{K}\sum_k W_{proto}h^c_k.
\]
In the training phase, the model is trained with the cross-entropy loss of the queries from $\mathcal{T}_{train}$:
%the classification probability of the $i$-th query feature $h_i$ are proportional to $\exp{s(z^c, h_i)}$.
%The loss for each task is Cross-Entropy Loss:
\[
L_{class} =-\frac{1}{M} \sum_i \sum_c y_i^c \log \frac{\exp{[s(z^c, h_i)]}}{\sum_{c'}\exp{[s(z^{c'}, h_i)]}},
\]
where 
$h_i$ is the $i$-th query feature, $s(z_c, h_{mm,i}) =z_c^\top W_{proto}h_{mm,i}$ is a score function containing $W_{proto}$ as a trainable parameter, and
$y_i^c \in \{0,1\}$ is the ground-truth label of the $i$-th query.
%The model parameters are trained with mini-batch of tasks sampled from $\mathcal{T}_{train}$.
%sample tasks  from $\mathcal{T}_{train}$ as mini-batch. 
In the test phase, the class prototypes are obtained from the supports, and the average score over the sampled tasks is reported.

\subsection{Few-Shot Image Classification with Language Description}
\begin{figure*}
		\includegraphics[width =150mm]{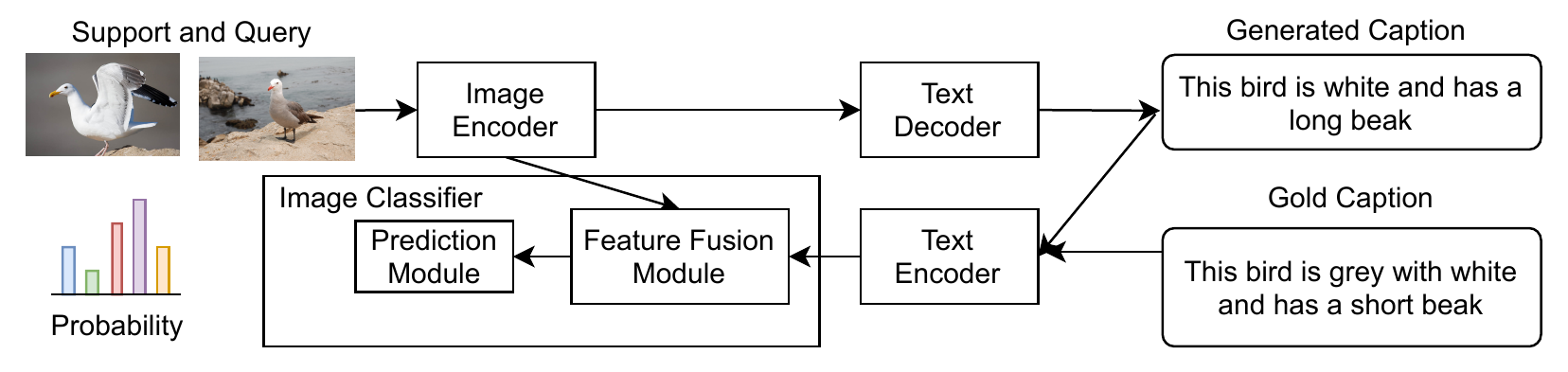}
		\caption{Overall structure of LIDE. The text decoder generates a caption on the basis of the image representation encoded by the image encoder. The text encoder obtains the text representation from the generated caption or a gold caption. The feature fusion module combines the image and text representation to generate the multi-modal representations. The prediction module outputs the classification probability. }
		\label{fig:model}
\end{figure*}

We focus on the few-shot image classification to verify the machine learning models' capability to learn new concepts from language descriptions in addition to the images. 
Previous studies have used the language description to improve the few-shot image classification.
The classifiers in the studies are based on ProtoNet, which corresponds to the model with the image encoder and prediction module in Figure \ref{fig:model}.
%but each overall model structure
%the proposed overall model structures are
%is different. %the representations fed to the classifiers are encoded by each proposed model.
\citet{lsl} proposed LSL by introducing a text decoder to ProtoNet to avoid overfitting by training the image encoder with a language generation loss. 
They observed that a text representation from a noisy machine-generated description was harmful for image classification. Accordingly, they viewed the text decoder as a regularizer and did not use a text encoder. 
RS-FSL 
\cite{rs_fsl} replaced the GRU \cite{gru} text decoder of LSL with a bi-directional transformer \cite{transformer}.

\citet{l3} were interested in describing the hidden states with natural language, but not in improving the image classification performance. 
%has the similar motivation to ours. %They explained the image hidden states by the natural language.
%They tackled few-shot image classification as one of the focused tasks. They were also interested in describing the hidden states by natural language, but  not in improving the image classification performance.
%They evaluate their model on few-shot image classification with language description. 
They proposed L3 by adding both a text decoder and encoder to ProtoNet.
They encoded an image into an image representation and decoded it into an explanation. The input of their image classifier was only the text representation encoded by the text encoder. 
L3 provided the explainability, but their model performance decreased.

%They were interested in describing the hidden states by the latent natural language, not in improving the image classification performance.

All of the aforementioned papers assumed that the image encoder was not pre-trained in a supervised or self-supervised fashion with external images. Our motivation %of this paper
is to clarify the benefit of the language description to learn novel visual concepts, and so we follow their setting. %We follows their setting.
%The classifiers in the papers are based on ProtoNet.

\section{Methods}
\subsection{Model}
We show the overall model structure of LIDE in Figure \ref{fig:model}. The model components are explained in the following.

\paragraph{Image Encoder}
We can use any network as the image encoder, and 
we used a 4-layer CNN as in the previous studies.
The output is the image feature $h_{img}$.

\paragraph{Text Decoder}
First, we map the image feature to the text feature space as follows:
\[
f_{I2T}(h_{img}) =\textrm{Linear}(\textrm{LayerNorm}(h_{img})),
\]
where LayerNorm is layer normalization \cite{layer_norm}.
%Then, we use a uni-directional 3-layer transformer for the text decoder.
Then, we pass the text feature vector to the text decoder as an encoder hidden state sequence of length 1. 
%We view the text feature as the encoder hidden states of length 1 although  the text feature is a vector, not a vector sequence. Therefore, the cross-attention module in the transformer decoder just adds the text feature vector multiplied by the layer-wise weight parameters to the decoder hidden states. We consider that this operation strengthen the awareness of the text feature derived from the image similarly to the style transfer.
The text decoder autoregressively generates the $j$-th token $t_j$ . The $j$-th token generation probability  $p_j$ is written as
\[
p_j =\Pr(t_j; f_{I2T}(h_{img}), t_{0:j-1}). 
\]
We used a uni-directional three-layer transformer.

\paragraph{Text Encoder}
We use BERT \cite{bert} for the text encoder, which outputs the last hidden states $H_{BERT}$.
The text feature $h_{text}$ is a weighted average pooling of $H_{BERT}$:
%, $\frac{1}{L}\sum h_{BERT, i}$, where $L$ is the text length.
%If the input text is the generated caption, $h_{text}$ is the weighted average pooling of the last hidden states, 
\[
h_{text} =\frac{1}{\sum p_jw_j}\sum p_j w_j h_{BERT, j},
\]
where the weight $p_j$ is the token generation probability in the text decoder. If a caption is user-generated, then $p_j$ is 1 for all tokens. The weight $w_j$ is 1 if the $j$-th token is not a stop word; otherwise, $w_j =0$.

The use of weighted average pooling with the text generation probability has two advantages.
First, the text encoder can ignore the low-confidence tokens. 
Second, the image classification loss back-propagates to the text decoder through the weight $p_j$. Because the discrete operation of text generation breaks the computation graph, we cannot back-propagate the gradient of $H_{BERT}$ to the decoder without the method. 
%Instead, we pass the gradient of $h_{text}$ through $p_j$, which enable the text decoder to consider the effects of the generated tokens on the image classification.

\paragraph{Feature Fusion Module}
The feature fusion module combines the single-modal features $h_{img}$ and $h_{text}$ into a multi-modal feature $h_{mm}$. 
Let $[ ; ]$ be the vector concatenations, $f_{T2I}$ be a mapping function from the text feature space to the image feature space, and $g$ be a linear function to $\mathbb{R}^2$.
$f_{T2I}$ is a three-layer %feedforward neural network (FFNN) with the rectified linear unit (ReLU) activation function.
FFNN with ReLU activation. 
The feature fusion operation is the weighted sum of the two features:
\iffalse
\begin{align}
%\[
\nonumber
[w_{img}; w_{text}] &=\textrm{softmax}(\textrm{Linear}( [h_{img}; h_{text}] )) \in \mathbb{R}^2, \\
%\]
%\[
\nonumber
h_{mm} &= w_{img}h_{img} + w_{text} f_{T2I}(h_{text}).
%\]
\end{align}
\fi
\[
[w_{img}; w_{lang}] =\textrm{softmax}(g( [h_{img}; h_{text}] )),
\]
\[
h_{mm} = w_{img}h_{img} + w_{text} f_{T2I}(h_{text}).
\]
%There are various methods to fusion features. 
%The advantage of this method is that it explicitly maps the features to the same feature space. 
%Our pilot experiments showed little difference within other fusion methods such as concat-and-linear transformation.

\paragraph{Prediction Module}
We use ProtoNet for the prediction module and replace $h$ with $h_{mm}$.
%The class prototype $z_c$ is the average of the $K$ support features $\frac{1}{K}\sum_{k}W_{proto}h_{mm,k}$, where $W_{proto}$ is is a trainable parameter.
%The logit representing that the $i$-th query $q$ is belong to the class $c$ is $s(z_c, h_{mm,i}) =z_c^\top W_{proto}h_{mm,i}$.

\subsection{Algorithms}
\paragraph{Loss Function} 
%We minimize \textcolor{blue}{a loss function} in the training phase to train the model.
For image classification, we compute two image classification losses: %$L_{class}$. 
%In each step, we clone the image feature.
$L_{class, gold}$ is computed from the image feature and the gold caption, while
$L_{class, gen}$ is computed from the image feature and the generated caption.
For text generation, we compute $L_{text}$ with  teacher-forcing and cross-entropy loss.

To enrich the mapped text feature $f_{T2I}(h_{text})$, we use the contrastive loss \cite{contrastive1, contrastive2} between the gold and generated captions. Let $v^c_{gold}$ and $v^c_{gen}$ be the averages of the mapped support text features in class $c$ from the gold and generated captions, respectively.
The similarity is $\cos (v_{gold}^{c \top} v^{c'}_{gen})$.
Then, the contrastive loss $L_{cntr}$ is
\begin{align*}
L_{cntr} &=-\frac{1}{2N}\sum_c 
\log \frac{\exp{[\cos (v_{gold}^{c \top} v^{c}_{gen})/\tau]}}
{\sum_{c'}\exp{[\cos (v_{gold}^{c \top} v^{c'}_{gen})/\tau]}} \\
%\textrm{logsoftmax}_{c'} \{\cos (v^c_{gold}^{\top} v^{c'}_{gen})/\tau \} \\
&-\frac{1}{2N}\sum_{c'} 
\log \frac{\exp{[\cos (v_{gold}^{c' \top} v^{c'}_{gen})/\tau]}}
{\sum_{c}\exp{[\cos (v_{gold}^{c \top} v^{c'}_{gen})/\tau]}}
,
\end{align*}
where $\tau$ is a temperature parameter.

The total loss that we minimize in the training phase to optimize the whole model parameters in an end-to-end manner is
\[
L =L_{class, gold} + L_{class, gen} + \lambda_{text} L_{text} + \lambda_{cntr} L_{cntr},
\]
where $\lambda_{text}$ and $\lambda_{cntr}$ are hyperparameters.

%\]

\paragraph{Pre-Training}
Following \citet{rs_fsl, fewshot_pre}, we pre-train the model with the training data for the downstream task. The pre-training consists of the standard image classification task, and we replace the prediction module with a linear classifier for all training classes. %Let $L_{class,pre}$ be Cross-Entropy Loss for the image classification with the gold captions.
The loss function is
$
%L =L_{class, pre} + \lambda_{text} L_{text}.
L =L_{class, gen} + \lambda_{text} L_{text}.
$
%\]

\paragraph{Caption Generation}
In the training phase, we use a greedy algorithm and random sampling for computational reasons.
In each step, we uniformly and randomly choose between the two algorithms. In random sampling, we restrict the candidate tokens to the top 20 tokens at each position.

In the test phase, we input the generated captions to the text encoder in the setting where the user-generated description is not available. The generation algorithm is beam search with a beam width of five and a length penalty of 0.5. 
%\iffalse
Thus, the token sequence $t_{1:l}$ is generated as
\[
\textrm{argmax}_{l, t_{1:l}} \frac{1}{l^{0.5}} \sum_{j=1}^l \log \Pr(t_j; f_{I2T}(h_{img}), t_{1:j-1}),
\]
%\fi
where $l$ is the text length.
The length penalty reduces the preference for words consisting of multiple subwords, such as `point \_y' (`pointy'). % to `long' or `short'.

\section{Evaluation}
\subsection{Dataset}
We used the Caltech-UCSD Birds (CUB) dataset \cite{cub} for evaluation.
It contains 200 bird species (classes) and 40-60 images for each class. The classes are split into 100 training classes, 50 development classes, and 50 test classes. 
We used this dataset for the $N =5$-way $K=1$-shot classification problem.
The number of query instances $M$ per class was 15. 

The CUB dataset has 10 captions for each image \cite{cub_caption}. %, which were created by Amazon Mechanical Turk (AMT) crowdworkers. 
For each step, we randomly sampled one caption from the 10 gold captions. % for each image.

\subsection{Metrics}
For the image classification, we report the average accuracy over 600 tasks, following the previous studies. To evaluate the generated caption quality, we used
%We evaluated the generated caption with 
major metrics for image captioning, $\textrm{BLEU}_\textrm{4}$ \cite{bleu}, METEOR \cite{meteor}, and $\textrm{ROUGE}_\textrm{L}$ \cite{rouge}. %, and CIDEr \cite{cider}.

\subsection{Implementation}
We pre-processed the images in the same way as \citet{lsl}.
The dimension of the image feature $h_{img}$ was 1600.
The text encoder and tokenizer were the pre-trained BERT-base-uncased model. 
The text encoder output dimension was 768.
The configuration of the transformer layers in the text decoder was the same as that of the T5-base decoder model \cite{t5}, but we did not use the pre-trained parameters for the text decoder because they did not contribute to the performance.
The parameter size of $W_{proto}$ was $1600 \times 1600$.
The other hyperparameters and optimization details are given in Appendix \ref{append:hyper}.

\subsection{Compared Models}
%We compared the previous models with 4-layer CNN image encoder.
ProroNet \cite{proto} was the baseline, with only the image encoder.
As for the other compared models, L3 \cite{l3} used a 200-dimensional GRU text encoder and decoder but did not use image representation for classification.
LSL \cite{lsl} used a 200-dimensional GRU text decoder for regularization.
RS-FSL \cite{rs_fsl} used a 2-layer, 768-dimensional, bi-directional transformer as the text decoder.
All models used a 4-layer CNN as the image encoder, along with the ProtoNet-based prediction module.

\subsection{Ablated Models}
%In order to
To evaluate the models that use a single-modal representation for image classification, we introduced the models ``No Text,'' ``No Image,'' and ``No Text Encoder'' by removing the feature fusion module from LIDE. %All models do not have the feature fusion module, and t
%They used only the image or text representations.
Each model corresponds to our implementation of the baseline models, as shown in Table \ref{tab:ablated_models}. %the results where the baseline models are implemented in the same 
%setting with our model. 
\iffalse
\begin{itemize}
\item The ``No Text'' model has the image encoder. The input of the classifier is only the image representation.
\item The ``No Image'' model has the image encoder, the text decoder, and the text encoder, but the image representation is not used for the classification. The input of the classifier is only the text representation.
\item The ``No Text Encoder'' model has the image encoder and the text decoder. The input of the classifier is the image representation.
\end{itemize}
The models correspond to our implementation of the baseline models. %the results where the baseline models are implemented in the same 
%setting with our model. 
\fi

%\iffalse
\begin{table}[t]
	\begin{center}
	\tabcolsep=4.5pt
		\scalebox{0.8}{
			\begin{tabular}{r|c|c|c|c}\hline
			{\small Model} & {\small Text Dec.} & {\small Text Enc.} & {\small Modal} & {\small Baseline} \\ \hline
			No Text & & & Image & ProtoNet \\
			No Image & \checkmark & \checkmark & Text & L3 \\
			No Text Enc. & \checkmark & & Image & LSL, RS-FSL \\
			\hline
		\end{tabular}}
		%\vspace{-0.5em}
		\caption{
		Setting of ablated single-modal models.} % Baseline column lists the corresponding baselines.}%\vspace{-0.5em}
		\label{tab:ablated_models}
		\end{center}
\end{table}
%\fi

%\subsection{Results}
\subsection{Evaluation Results for LIDE as Few-Shot Image Classification Model}
\subsubsection{Main Results}
\paragraph{Performance with Machine-generated Description}
Table \ref{tab:main} summarizes the results.
LIDE outperformed
%Compared to 
ProtoNet, LSL, and RS-FSL, which use an image classifier using image representations only.
%, % the models using image representation, 
%we found that 
%\textbf{
%the fusion of the multi-modal representations improved the performance 
% \textcolor{blue}{even when our model used a noisy prediction of the description}. 
%Note that the bi-directionality of the RS-FSL text decoder can be applied to LIDE, and this would improve its performance.
%Compared to 
%Although L3 that uses text representation for image classification performed poorly, LIDE was able to improve the performance.
%Among the models using text representation,
%\textbf{
%To our knowledge, this is the first study that achieved the improvement of few-shot image classification performance with the mechanism of encoding natural language description in the test time.
%}. 
%\textcolor{red}{
%As for comparison to L3, %\textcolor{blue}{with the model using text encoder},
LIDE improved the image classification performance with the mechanism of encoding text and combining multi-modal representations, while L3, which uses an image classifier with text representations only, harmed their performance.
%This 
%\textcolor{magenta}{The new techniques LIDE introduced enabled the improvements by using the text representaions}.
%, and} \textcolor{magenta}{LIDE's} text encoding mechanism is essential to allow the user input in the test phase. %However, the previous models did not succeed in introducing it without harming the performance.

\begin{table}[t]
	\begin{center}
	\tabcolsep=4.5pt
		\scalebox{0.8}{
			\begin{tabular}{r|c|c|c|c|c}\hline
			{\small Model} & {\small Accuracy} & {\small Img. Enc.} & {\small Text Enc.} & {\small Fusion} & {\small Text Dec.} \\ \hline
			ProtoNet & 57.97 $\pm$ 0.96 &
			\checkmark & & &  \\
			L3 & 53.96 $\pm$ 1.06 &
			 & \checkmark & & \checkmark \\
			LSL & 61.24 $\pm$ 0.96 &
			\checkmark & & & \checkmark \\
			RS-FSL & 65.66 $\pm$ 0.90 &
			\checkmark & & & \checkmark \\
			LIDE & \textbf{67.53} $\pm$ 0.91 &
			\checkmark & \checkmark & \checkmark & \checkmark \\\hline
		\end{tabular}}
		%\vspace{-0.5em}
		\caption{
		Performance of the compared models.}%\vspace{-0.5em}
		\label{tab:main}
		\end{center}
\end{table}

\iffalse
\begin{table}[t]
	\begin{center}
		\scalebox{0.75}{
					\tabcolsep=4.5pt
	\begin{tabular}{r|c|c|c|c|c|c}\hline
			 & \multicolumn{2}{c|}{{\small Accuracy}} & {\small Img.} & {\small Text} & & {\small Text} \\ %\hline
						{\small Model} & {\small CUB} & {\small Shape} & {\small Enc.} & {\small Enc.} & {\small Fusion} & {\small Dec.} \\ \hline
			ProtoNet & 57.97 $\pm$ 0.96 & 00.00 $\pm$ 0.00 &
			\checkmark & & &  \\
			L3 & 53.96 $\pm$ 1.06 & 00.00 $\pm$ 0.00 &
			 & \checkmark & & \checkmark \\
			LSL & 61.24 $\pm$ 0.96 &  00.00 $\pm$ 0.00 &
			\checkmark & & & \checkmark \\
			RS-FSL & 65.66 $\pm$ 0.90& 00.00 $\pm$ 0.00 &
			\checkmark & & & \checkmark \\
			LIDE & 67.53 $\pm$ 0.91 & 00.00 $\pm$ 0.00 &
			\checkmark & \checkmark & \checkmark & \checkmark \\
			\hline
		\end{tabular}}
		%\vspace{-0.5em}
		\caption{
		Performance of compared models on CUB and shapeworld datasets.}%\vspace{-0.5em}
		\label{tab:main}
		\end{center}
\end{table}
\fi

\paragraph{Performance with User's Description}
% future directionのところにまとめる？
%いや，大事な貢献だから2番目に書いておく
The advantage of the text encoder is that it enables textual input by users.
%the availability of textual input by users.
Specifically, a user can use language as an explanation from humans to machines 
by feeding
%to feed 
the textual features captured from the input image to the model. % to the model.
%can explain the textual features captured from the input image with language to the model.
In addition, if a user objects to the model's explanations, the user can edit them to %obtain the expected result.
correct the model's misunderstanding.

To evaluate LIDE in this setting, we viewed the gold captions as user descriptions. 
%\textcolor{red}{If the model outputted the wrong result, we encoded one of the gold captions and obtained new prediction.}
As the CUB dataset has 10 gold captions per image, we selected one gold caption and fed it to the model to maximize the similarity to a generated caption and thus simulate a user editing the generated caption. The similarity was defined as the bi-gram precision of the gold caption with respect to the generated caption.
%Therefore, we investigate the performance in the setting where the gold caption is available to evaluate the performance in the setting where the user can add the description of the image.

Table~\ref{tab:gold} shows
%lists the results, which show 
that 
%the language information improved the performance, and 
%\textbf {
the performance was improved significantly when a high-quality gold caption was given.
%the user input a high-quality description. 
%} %the caption desctibing the charactaristic of the bird.
%\textcolor{blue}{This improvement was brought by the text encoder and the feature fusion module in LIDE, which enable to use text representations.}
%\textcolor{blue}{This is the first research on using both vision and language representation for few-shot image classification. 
%\textcolor{red}{
%The ability to use the multi-modal representations is %}
%essential for the model to reflect a user's description.}
%In addition, 
However, the performance declined when a wrong caption was given, which was a randomly sampled caption from all captions in  $\mathcal{T}_{test}$.
%we passed the randomly sampled caption from all captions in $\mathcal{T}_{test}$. Then, the model performance declined.
We conclude that the model's output depends on the quality of the description.

\begin{table}[t]
	\begin{center}
		\scalebox{0.8}{
			\begin{tabular}{r|c}\hline
			%No Description ($w_{lang} =0$) &  58.88 \pm 1.11  \\
			Random Description & 58.89 $\pm$ 0.93 \\
			Generated Description &  67.53 $\pm$ 0.91 \\
			Gold Description &  \textbf{73.08} $\pm$ 0.88  \\
			\hline
		\end{tabular}}
		%\vspace{-0.5em}
		\caption{
		Performance with each kind of description.}%\vspace{-0.5em}
		\label{tab:gold}
		\end{center}
\end{table}

\subsubsection{Ablation Study}
%Table~\ref{tab:ablation} shows the ablation study.

\paragraph{Evaluation on modalities}
The first set of rows in Table~\ref{tab:ablation} lists ablation study results for the different modalities that LIDE uses. %These results also correspond to the results when the other models were implemented in the same setting as our model. 
We confirmed that LIDE using all the modalities outperformed the compared models using part of the modalities. %We conclude that 
The image and text representations complemented each other, as will be discussed later. %in Sec \ref{sssec:attribution}.
%\textcolor{magenta}{We also observed that the performance of the baseline models improved except for the ``No Image'' model, mainly because of their hyperparameter settings. Training of the ``No Image' model is unstable due to its transformer architecture.}

\paragraph{Evaluation on introduced techniques}
The second set of rows in Table~\ref{tab:ablation} lists the techniques that were introduced in LIDE:
image classification loss with generated captions, contrastive learning, weighted average pooling, and random sampling during caption generation.
%compared LIDE to the ablated models. 
We confirmed that all of these techniques contributed to the performance of LIDE. 

We assume that the loss with the generated captions decreases the discrepancy between the train and test phases.
The contrastive loss enriches the text representation and makes the representations of the generated text and the gold caption close. The weighted average pooling reduces the effect of noisy generated text. The random sampling in the training phase contributes to the generation of diverse captions.

\paragraph{Evaluation on pre-training of text encoder}
%Surprisingly, 
We also found that the multi-modal feature was useful even when the text encoder was trained from scratch. 
The text representation could assist in the image representation without BERT pre-training, because the captions in the CUB dataset are restricted to the descriptions of birds, and the training data thus covers the space of the captions well.
%Also, the weighted average pooling improved the robustness for the unseen captions.

\begin{table}[t]
	\begin{center}
		\scalebox{0.8}{
			\begin{tabular}{r|c}\hline
			LIDE &  \textbf{67.53} $\pm$ 0.91  \\ \hline
			No Text (ProtoNet) &  62.22 $\pm$ 0.92  \\
			No Image (L3) &  49.60 $\pm$ 1.03  \\
			No Text Encoder (LSL, RS-FSL) & 63.10 $\pm$ 0.90  \\\hline
			No $L_{class,gen}$ &  61.96 $\pm$ 0.90  \\ 
			No Contrastive Loss &  64.60 $\pm$ 0.88  \\  
			No Weighted Average Pooling &  66.16 $\pm$ 0.93  \\
			No Random Sampling &  66.40 $\pm$ 0.93  \\ 
			None of the Above  &  59.68 $\pm$ 0.98  \\ \hline
			No BERT Pre-Training &  66.42 $\pm$ 0.94  \\  \hline

		\end{tabular}}
		%\vspace{-0.5em}
		\caption{
		Ablation study results.}%\vspace{-0.5em}
		\label{tab:ablation}
		\end{center}
\end{table}

\iffalse
\subsubsection{Evaluation for non-novel concepts}
In addition, we compared the models in the setting where the image encoder can be pre-trained with the supervised ImageNet dataset \cite{imagenet}. This experiment is out of our main focus because we are interested in the benefit of the language description to learn the novel concept, following to the traditional setting of the research. The supervised training with the ImageNet dataset teaches many visual concepts in the test set of the CUB dataset. Through this evaluation, we verify the benefit of the language description to enrich the representation of the known image concept.

Table \ref{tab:pre-trained} shows the results.
We used 50-layer ResNet \cite{resnet} for image encoder.

\begin{table}[t]
	\begin{center}
		\scalebox{0.75}{
			\begin{tabular}{r|c}\hline
			LIDE &  81.86 \pm 0.80  \\
			+ Gold Caption &  82.86 \pm 0.83  \\ \hline
			No Text (ProtoNet) &  82.80 \pm 0.80  \\
			No Image Encoder (L3) &  XX.XX \pm 0.XX  \\
			No Text Decoder (LSL, RS-FSL) &  83.72 \pm 0.76  \\
			\hline
		\end{tabular}}
		%\vspace{-0.5em}
		\caption{
		Performance with pre-trained CNN encoder. }
		%\vspace{-0.5em}
		\label{tab:pre-trained}
		\end{center}
\end{table}
\fi

%\subsection{Proposed Model as Interpretable Machine Learning Model}
\subsection{Evaluation Results for LIDE as Interpretable Machine Learning Model}

%\subsubsection{Generated Caption as Explanation}

%We investigated whether the quality of the generated caption has effect on the classification performance.

\paragraph{Evaluation on quality of generated captions}
First, generated captions are insufficient as explanations if they are not accurate. 
%We evaluated the generated caption with major metrics of image captioning, $\textrm{BLEU}_\textrm{4}$ \cite{bleu}, METEOR \cite{meteor}, , and $\textrm{ROUGE}_\textrm{L}$ \cite{rouge}. %, and CIDEr \cite{cider}.
The upper bound was the CNN-LSTM model pre-trained with the MSCOCO \cite{mscoco} dataset from \citet{cub_ub}. Although we had no training data without the CUB dataset, Table \ref{tab:corr} shows that the differences between LIDE and the upper bound were only 2.0 points for METEOR and 3.3 points for ROUGE. 
%CIDEr considers the TF-IDF score in the dataset. Rare words in the CUB dataset are assessed heavily. However, the model trained only with CUB dataset generates rare words less than the pre-trained model does, due to the over-fitting. Therefore, the CIDEr score of LIDE is low.

\paragraph{Evaluation on consistency between generated captions and image classification}

%\citet{lsl} reported that an ablated language supervision study by shuffling the captions across tasks, and training with wrong image-caption pairs did not worsen the classification accuracy significantly comapred to training with correct image-caption pairs.

%生成したキャプションが正しい・誤りを持つときは分類結果も正しい・誤ることがモデルの説明と分類結果の一貫性の観点からは望まし
When the generated captions are correct (respectively, incorrect), the classification results should also be correct (incorrect) in terms of the consistency between explanations and classification results.

Accordingly, we calculated the Spearman rank-order correlation coefficient between the captioning scores and the classification scores. We divided 2,953 test-split images into 30 bins in accordance with the ascending order of each captioning score, and we computed the average image classification accuracy in each bin.

Table \ref{tab:corr} lists the results. 
We confirmed a positive correlation between the quality of the generated captions and the prediction accuracy of LIDE. However, LIDE without the text encoder, which corresponds to LSL~\cite{lsl} and RS-FSL~\cite{rs_fsl}, showed a low correlation. This result demonstrated the importance of the text encoder and the feature fusion modules introduced in LIDE.
%Therefore, the wrong caption can be the evidence of the wrong classification, and user may obtain improved prediction by modifying the generated captions.
%\citet{lsl} reported that an ablated language supervision study with LSL in which shuffling of captions across classes did not significantly degrade the image classification accuracy, and the results demonstrated low consistency between the generated captions and the image classification by LSL.

\begin{table}[t]
	\begin{center}
		\scalebox{0.8}{
			\begin{tabular}{r|c|c|c}\hline
			  & BLEU$_\mathrm{4}$ & METEOR & ROUGE$_\mathrm{L}$ \\ \hline
			 UB: Caption & 59.0 & 36.1 & 69.7  \\ \hline
			 No Text Enc.: Caption & 50.0 & 34.6 & 67.2  \\
			 Correlation & 0.114 & 0.201 & 0.217 \\ \hline
			 LIDE: Caption & 48.1 & 34.1 & 66.4  \\
			 Correlation & 0.309$^\dag$ & 0.468$^*$ & 0.436$^{**}$ \\
			\hline
		\end{tabular}}
		%\vspace{-0.5em}
		\caption{
		Captioning scores and correlations to the prediction scores.\dag :~$p <0.1$, * :~$p <0.05$, ** :~$p <0.01$}%\vspace{-0.5em}
		\label{tab:corr}
		\end{center}
\end{table}

\begin{figure}
		\includegraphics[width =75mm]{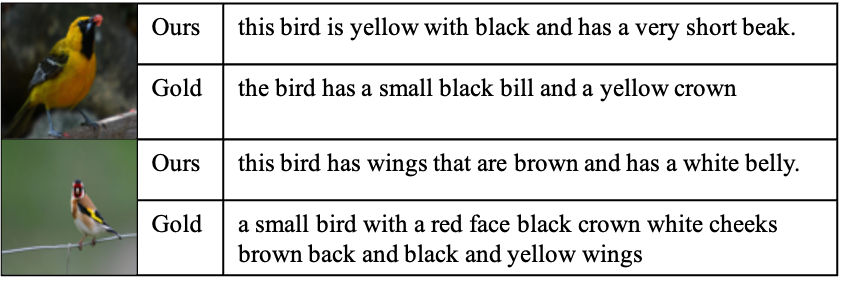}
		\caption{Examples of generated captions and gold captions.}
		\label{fig:caption}
\end{figure}

\paragraph{Qualitative analysis on generated captions}
Figure \ref{fig:caption} shows examples of the generated captions. 
We found that the captions captured the birds' characteristics. However, the structures of the captions were uniform, and they could not describe a birds' most distinctive element, such as the red face in the second example. 
We believe that overfitting to the $5$-class classification problems with the small dataset
%CUB training data 
caused this problem.

We restricted the image encoder to a 4-layer CNN for fair comparison to the existing models, and we did not use external training data to validate the ability to learn novel classes from language descriptions. Removal of the limitations would improve the performance of LIDE as an explainable machine learning model. 

\subsection{Discussion}
In this section, we clarify four reasons why text representations are useful in the few-shot image classification task. 
Specifically, we compared the multi-modal feature space of LIDE with the gold captions to the image feature space of the ``No Text'' and ``No Text Encoder'' models. %LIDE used the gold caption.% with the highest prediction score without considering the correctness of the predictions.

%\subsubsection{Distribution of the Feature Representations}

%\paragraph{Inner/inter class distance in multi-modal (fusioned) feature spaces}
\paragraph{How are classes distributed in each modal feature space?}
First, we calculated the inner- and inter-class distances in each feature space. %\textcolor{red}{three features}.
Table \ref{tab:distribution} lists the results.
%\textcolor{blue}{The inner-class distances were similar among the three models. However, the inter-class distances were different between the image representations of the ``No Text'' and ``No Text Encoder'' models and the multi-modal representation of LIDE.}
The inner-class distances in the multi-modal feature space of LIDE were smaller than those in the image feature space; the inter-class distances of LIDE were larger. As a result, the clusters %of multi-modal features corresponding to each class 
were distributed well in the multi-modal feature space.
%As a result, the distribution of the features in each class di not depend on the modals. However, 
%\textbf{
%the clusters of multi-modal features corresponding to each class were further apart than those of the image features.
%}

We believe that this is because the captions describe the similarities and differences between the images more obviously than the images themselves do. % We take the examples as the first and second images in Figure \ref{fig:caption}. 
For example, to determine that two birds belong to different species, one piece of evidence is the belly color. 
The captions can explain this information clearly, e.g., ``yellow belly'' and ``white belly''. 
From the image, however, the extraction of this information requires multiple steps such as locating the belly and specifying its color.

%\paragraph{Dimension size of latent feature space}
\paragraph{What are characteristics of latent feature spaces?}
Second, we examined the dimensions of the feature spaces. 
\citet{lid_feature} observed that the features embedded in a manifold with a smaller latent dimension are more generalized. They evaluated the latent dimension by using the average of the local intrinsic dimensions (LID) of the features.
The LID measures the number of dimensions of a feature manifold in the neighbor of $x$, and it can be estimated as
\[
\hat{\textrm{LID}}(x) = - \left\{\frac{1}{n_{nn}} \sum^{n_{nn}}_{i}\log\left(\frac{r_i(x)}{r_{n_{nn}}(x)}\right) \right\}^{-1}
\]
by maximum likelihood estimation, where $n_{nn}$ is the number of the nearest neighbors and $r_i$ is the Euclidean distance from $x$ to the $i$-th nearest neighbor \cite{lid_mle1, lid_mle2}. We set $n_{nn} =20$ in accordance with the previous studies.

\begin{figure}
		\includegraphics[width =75mm]{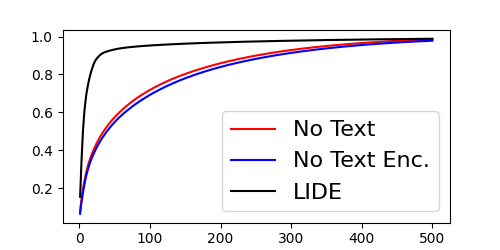}
		\caption{Cumulative contribution rates for PCA of each representations.}
		\label{fig:pca}
\end{figure}

Table \ref{tab:distribution} lists the estimated LIDs of the features of each model. In addition, we applied principal component analysis (PCA) to the features, and Figure \ref{fig:pca} shows the cumulative contribution rates. All features were embedded in $\mathbb{R}^{1600}$.

%Surprisingly, %\textbf{
The multi-modal features existed in a manifold with a smaller latent dimension than those of the image-only representations.
%}
%Both results show that 
Therefore, the text representation contributed to shrinking the representation manifold to a smaller dimension.
We assume that the text controlled the main focus among the many objects in an image.
%There are a lot of objects in the image, and there are many potential captions for the image.
For example, the captions in the CUB dataset describe the characteristics of birds.
As a result, the model can extract the important information from captions for the downstream image classification task.
In contrast, an image has much information, such as the background,
%and the direction of the bird. Therefore, the image feature contains the information not required for the classification of bird species
and image features thus require a larger-dimension manifold.

% distribution
\begin{table}[t]
	\begin{center}
		\scalebox{0.8}{
			\begin{tabular}{r|S|S|S}\hline
			 & {\small Inner-Class Dist.} & {\small Inter-Class Dist.} & {\small LID} \\ \hline
			{\small No Text (Image)} & 0.504 & 0.592 & 17.8 \\
		    {\small No Text Enc. (Image)} &  0.526 & 0.609 & 19.0 \\
			{\small LIDE (Fusion)} & 0.459 & 0.709 & 6.73 \\
			\hline
		\end{tabular}}
		%\vspace{-0.5em}
		\caption{
		Distribution of feature representations. }%\vspace{-0.5em}
		\label{tab:distribution}
		\end{center}
\end{table}

%\subsubsection{Robustness for Noisy Inputs}
\paragraph{Are multi-modal representations robust for noisy images?}
Next, we hypothesize that language descriptions can help to classify noisy or obscure images.
To verify our hypothesis, we performed experiments in two heuristic noisy settings:
One consisted of grayscale images and the other consisted of images that were adversarially attacked via the fast gradient sign method \cite{fgsm}.
We compared LIDE with gold captions to the ablated models using the image representations and evaluated their performance with the noisy images in the test phase.

Table \ref{tab:noisy} lists the result.
When a caption was not provided, the classification accuracy dropped greatly in both settings. However, LIDE reduced the decline by virtue of the textual information.
These results indicate that %\textbf{
the text representation may be useful for classifying certain ill-conditioned images, %}, 
such as an image in which the bird is extremely small.

\begin{table}[t]
	\begin{center}
		\scalebox{0.8}{
			\begin{tabular}{r|c|c|c}\hline
			 & {\small Original} & {\small Grayscale} & {\small Adversarial} \\ \hline
			{\small No Text} &  62.22 $\pm$ 0.92  & 38.04 $\pm$ 0.78 & 33.09 $\pm$ 0.69 \\
			%No Image &  62.00 \pm 1.15 & 62.00 \pm 1.15  & 62.00 \pm 1.15 \\
			%No Image &  49.60 \pm 1.03 & 33.61 \pm 0.79  & 26.49 \pm 0.62 \\
			{\small No Text Enc.} &  63.10 $\pm$ 0.90 & 38.55 $\pm$ 0.80 & 31.30 $\pm$ 0.65 \\
			%LIDE &  78.09 \pm 0.79  & 30.75 \pm 0.58 & 42.66 \pm 0.85 \\ gyaku
			{\small LIDE} &  78.09 $\pm$ 0.79  & 59.81 $\pm$ 0.97 & 56.31 $\pm$ 1.01 \\
			\hline
		\end{tabular}}
		%\vspace{-0.5em}
		\caption{
		Performance in noisy settings.}%\vspace{-0.5em}
		\label{tab:noisy}
		\end{center}
\end{table}

%\subsubsection{Attributions Recovered from Representations}
\paragraph{What information do text representations have?}
\label{sssec:attribution}
%Last, we investigated what information the image and the text representations have. Similar to the previous studies in natural language processing that discovers linguistic properties \cite{probing1, probing2} in the text representations, we performed the probing test to the each modal representation.
Finally, we performed a probing test for each modality of representation, as in previous natural language processing (NLP) studies that discovered linguistic properties \cite{probing1, probing2} in the text representations.
The CUB dataset has annotations of the birds' attributions, and we recover the attribution labels from $h_{img}$ and $h_{text}$. Each image has $\{0,1\}$ labels for 312 attributions. %We tackled the probing task to recover the attribution label from $h_{img}$ and $h_{text}$.

We used our trained model to obtain $h_{img}$ and $h_{text}$ for all images and captions. Then, we used linear classifiers $W_{img,attr} \in \mathbb{R}^{1600 \times 312}$ and $W_{lang,attr} \in \mathbb{R}^{768 \times 312}$, which were trained the linear classifiers with binary-cross-entropy loss in the training split. 
Next, we determined the thresholds 
%for each attribution with sigmoid activation 
in the development split and obtained prediction results in the test split. %.
%Third, we obtained the prediction results of each image and caption for all attributions. 
Finally, we performed a Wilcoxon signed-rank test between the image results and the text results.

%We sampled one caption for each image in the development and test splits. If the attribution was not described in the caption although the label was 1, this example was ignored.
%We removed the attributions on the following condition: The number of correct prediction is less than 20. The number of positive example is less than 20. %Positive example means that label is 1 and the attribution is described in the caption.
%Finally, 68 attributions were remained.

Table \ref{tab:test} lists the numbers of attributions having significance at a $p$-value of 0.05. Among the 68 attributions, 60 were recovered more easily from one modality than from the other modality. In other words, 
%\textbf{
the image and text representations complemented each other. 
%}.
Most of the attributes favored the image representation, but 17 of them favored the text representation.
Table \ref{tab:attribution} lists those 17 attributes, and we can observe two main characteristics among them.
First, the colors black and white were recovered from the text representation. These attributes may be difficult to recover from an image because of light and shadow.
The second characteristic was a spotted pattern, which is obscure in an image with $84\times84$ pixels.

\begin{table}[t]
	\begin{center}
		\scalebox{0.8}{
			\begin{tabular}{c|c|c}\hline
			 Image & Text & No Significance \\ \hline
			 43 & 17 & 8 \\
			\hline
		\end{tabular}}
		%\vspace{-0.5em}
		\caption{
		Numbers of significant attributions.}%\vspace{-0.5em}
		\label{tab:test}
		\end{center}
\end{table}

%69属性のうち，62属性で有意差があった
%よって，相補的である
%そのうちimageが有意なのは60個
%languageが有意なのは20個
%20個全部列挙する？
\begin{table}[t]
	\begin{center}
		\scalebox{0.9}{
			\begin{tabular}{c}\hline
\Verb|has_bill_shape::all-purpose| \\
\Verb|has_wing_color::white| \\
\Verb|has_back_color::black| \\
\Verb|has_breast_color::white| \\
\Verb|has_throat_color::white| \\
\Verb|has_eye_color::brown| \\
\Verb|has_eye_color::white| \\
\Verb|has_nape_color::black| \\
\Verb|has_nape_color::white| \\
\Verb|has_nape_color::red| \\
\Verb|has_belly_color::white| \\
\Verb|has_size::small_(5_-_9_in)| \\
\Verb|has_back_pattern::spotted| \\
\Verb|has_tail_pattern::spotted| \\
\Verb|has_belly_pattern::spotted| \\
\Verb|has_crown_color::grey| \\
\Verb|has_crown_color::black| \\
			\hline
		\end{tabular}}
		%\vspace{-0.5em}
		\caption{
		Attributions that were significantly recovered from text representations.}%\vspace{-0.5em}
		\label{tab:attribution}
		\end{center}
\end{table}

\section{Related Work}
%few-shot
%few-shot image classificationの解き方3流派について触れる？->No

%test timeにもgold langが使える設定ではいろいろな研究がある．（lsl論文の4本）
%使えないのでgenerationにだけ使う研究は2本．
%この2つを組み合わせた研究と位置付けられる
%その他の研究にも2本あるけど，ここではいいかー

\paragraph{Image classification with language}
%Although our setting does not provide the gold captions in the test time, 
Several studies have provides gold language information in the test phase.
For zero-shot learning or few-shot learning, class-label words are used as additional information \cite{NIPS2013_7cce53cf,socher2013zero,xing2019adaptive}.
Moreover, \citet{he2017fine,alice} used language descriptions for the standard image classification problem, in which the classes in the training and test phases are the same. 
The few-shot image classification is a more challenging problem that requires the capability of learning novel concepts from language descriptions.
%and thus did not validate the capability to learn novel concepts in unseen classes from the language.}
%Pre-trained language models also use them as the human-created prompt in natural language processing \cite{gpt-2, gpt-3}.
%Our framework joint two paradigm of the above papers and the papers without the gold caption in the test time by using decode-and-encode model.

%grounding
%最終的にどうまとめたいのかもう少し考える
%\paragraph{Visual grounding in vision and language tasks}
\iffalse
\textcolor{magenta}{
Visual grounding is a capability required for vision and language tasks such as image captioning \cite{caption} and visual question answering (VQA) \cite{vqa}. 
Previous works were motivated to acquire this ability in image captioning \cite{grounded_caption1, grounded_caption2, grounded_caption3, grounded_caption4} and VQA \cite{grounded_vqa1, grounded_vqa2}.
However, there has been discussion 
on the extent to which visual grounding has been achieved.
Although \citet{grounded_vqa1} used a human attention map to improve the visual grounding in VQA, 
\citet{negative_vqa} demonstrated 
that the improvement resulted not from the visual grounding itself, but the regularization effect of a human attention map.
The report in \citet{negative_vqa} is consistent with
%corresponds to 
the observation in \citet{lsl} that language information is useful only for regularization.
In contrast, the improved performance by LIDE and our findings suggest that the text representation is useful beyond the regularization effect.}
% results in this paper raise new arguments on this point.}
\fi

\paragraph{Textual explanation of image representation}
Explainability of artificial intelligence (XAI) has attracted much attention \cite{blackbox_nlp}. Papers have proposed methods to generate an image description for XAI in the image classification task and visual question answering (VQA) task \cite{hendricks2016generating, hendricks2018grounding, li2018tell}. 
In contrast to those studies, the motivation of this paper is to decode and encode such descriptions to improve the few-shot image classification performance. 
%\textcolor{blue}{Accordingly, we restricted the backbone image encoder for fair comparison,} but we could integrate the findings of XAI studies into LIDE.
We could integrate the findings of XAI studies into LIDE.
%, but we restricted the backbone image encoder in this paper for fair comparison.}

\paragraph{Analysis of image and text representations}
Previous papers have investigated why language information is useful in vision and language tasks.
\citet{mm_anal1} also found that the image and text representations complemented each other; for example, taxonomic attributes are captured well in the language.
\citet{mm_anal2} observed that the attention heads of the multi-modal pre-trained models ground elements of language to image regions.

\section{Conclusion}
We tackled the few-shot image classification task through learning of novel concepts from language descriptions of images. We observed that machine- and user-generated descriptions improved the few-shot image classification performance.
We also found that the generated captions explained the input image and were consistent with the prediction performance.

Our experiments also revealed four reasons why the text representation improved the performance:
the inner-class distances of the multi-modal representations are smaller and the inter-class distances are larger than those of image representations;
multi-modal representations are embedded in a space with a smaller latent dimension;
multi-modal representations are robust for noisy images;
and certain 
types of knowledge are easily recovered
from text representations.

\iffalse
\textcolor{blue}{Recently, multi-modal pre-trained models have been published \cite{vilbert, clip}.
We believe that our findings provide worthwhile insights for analysis of such multi-modal pre-training methods.}
\fi

Humans can learn concepts from language and explain them with language, but this is still difficult for machine learning models.
This study sheds light on the importance of interactivity in explanating with language in machine learning.
%%%We believe that this paper contributes to the %development of AI toward the 
%%%advent of the convivial society, in which AI learns and explains like human.

\bibliography{theme}
\bibliographystyle{acl_natbib}

\appendix

\section{Experimental Setup}
\label{append:hyper}
We trained all the models on an NVIDIA Quadro RTX 8000 (48GB), and each experiment took almost one day.
The hyperparameter settings are listed in Table \ref{tab:hyper1}. 
We used the Adam optimizer~\cite{adam}, PyTorch~\cite{pytorch}, and transformers~\cite{transformers}.
Stop words were implemented with NLTK~\cite{nltk}, and ``bird'' was added to the stop words.
The training of the ``No Image'' ablated model was unstable due to the transformer architecture, so we used greedy decoding in the test phase to reduce the discrepancy between the train and test phases.

\begin{table}[ht!]
\begin{center}
    \scalebox{0.8}{
		\begin{tabular}{c|c|c}\hline
                & Pre-Training & Fine-Tuning \\ \hline
				 {\small Batch size} & 128 & 100 \\
				 {\small \# Epoch}s & 100 & 1500 \\
				 {\small Learning rate for main model} & 1e-3 &  1e-3 \\
				 {\small Learning rate for text encoder} & 1e-3 &  1e-4 \\
				 {\small Learning rate for text decoder} & 1e-5 &  1e-5 \\
				 {\small $\lambda_{text}$} & --- & 10 \\
				 {\small $\lambda_{cntr}$} & --- & 0.1 \\
				 {\small $\tau$} & --- & 0.05 \\
				 \hline
				\end{tabular}}
	\caption{Hyperparameters.}
	\label{tab:hyper1}
\end{center}
\end{table}

%attribution実験
For the experiments on recovering attributions (Section \ref{sssec:attribution}),
we used sigmoid activation and determined the thresholds for each attribution.
We sampled one caption for each image in the development and test splits. When an attribution was not described in a caption even though the label was 1, the example was ignored.
We also removed attributions when the number of correct predictions was less than 20 or the number of positive examples was less than 20. 
Here, a positive example means one for which the label is 1 and the attribution is described in the caption.
As a result, 68 attributions remained.

\end{document}